\documentclass{article}

\usepackage{arxiv}

\usepackage[utf8]{inputenc} % allow utf-8 input
\usepackage[T1]{fontenc}    % use 8-bit T1 fonts
\usepackage{hyperref}       % hyperlinks
\usepackage{url}            % simple URL typesetting
\usepackage{booktabs}       % professional-quality tables
\usepackage{amsfonts}       % blackboard math symbols
\usepackage{nicefrac}       % compact symbols for 1/2, etc.
\usepackage{microtype}      % microtypography
\usepackage{lipsum}
\usepackage{graphicx}
\usepackage{booktabs}
\usepackage{todonotes}
\setuptodonotes{inline}

\usepackage{amsmath} 
\usepackage{amssymb} 
\usepackage{stmaryrd}
\usepackage{subcaption}
\usepackage{tabularx}
\usepackage{comment}
\usepackage{pgfplots}
\usepackage{placeins}
\usepackage[authoryear]{natbib}
\renewcommand{\cite}[1]{\citep{#1}}

\title{A Flexible Class of Dependence-aware \\ Multi-Label Loss Functions}

\author{
  Eyke H{\"u}llermeier\\
  Paderborn University\\
  Paderborn, Germany \\
  \texttt{eyke@upb.de} \\
   \And
  Marcel Wever\\
  Paderborn University\\
  Paderborn, Germany \\
  \texttt{marcel.wever@upb.de} \\
   \And
  Eneldo Loza Mencia\\
  Technical University Darmstadt\\
  Darmstadt, Germany \\
  \texttt{research@eneldo.net}\\
  \And
   \And
  Johannes Fürnkranz\\
  Johannes Kepler University\\
  Linz, Austria \\
  \texttt{juffi@faw.jku.at}\\
   \And
  Michael Rapp\\
  Technical University Darmstadt\\
  Darmstadt, Germany \\
  \texttt{mrapp@ke.tu-darmstadt.de}\\
}

\newcommand{\moebius}{m}
\renewcommand{\vec}[1]{\boldsymbol{#1}}

\newcommand{\given}{\, | \,}

\newcommand{\Prob}{P}
\newcommand{\prob}{p}

\newcommand{\cX}{\mathcal{X}}
\newcommand{\cL}{\mathcal{L}}
\newcommand{\cY}{\mathcal{Y}}

\newcommand{\cD}{\mathcal{D}}
\newcommand{\bx}{\boldsymbol{x}}
\newcommand{\by}{\boldsymbol{y}}

\newcommand{\bY}{\mathbf{Y}}
\newcommand{\bh}{\boldsymbol{h}}
\newcommand{\bs}{\boldsymbol{s}}

\newcommand{\loss}{\ell}

\newcommand{\fromto}{\longrightarrow}

\pgfplotsset{every axis plot/.append style={densely dashed,very thick}}

\definecolor{color1}{HTML}{023EFF}
\definecolor{color2}{HTML}{FF7C00}
\definecolor{color3}{HTML}{1AC938}
\definecolor{color4}{HTML}{E8000B}
\definecolor{color5}{HTML}{8B2BE2}
\definecolor{color6}{HTML}{9F4800}
\definecolor{color7}{HTML}{F14CC1}
\definecolor{color8}{HTML}{FFC400}
\definecolor{color9}{HTML}{00D7FF}
\definecolor{color10}{HTML}{A3A3A3}

\begin{document}
\maketitle

\begin{abstract}
Multi-label classification is the task of assigning a subset of labels to a given query instance. For evaluating such predictions, the set of predicted labels needs to be compared to the ground-truth label set associated with that instance, and various loss functions have been proposed for this purpose. In addition to assessing predictive accuracy, a key concern in this regard is to foster and to analyze a learner's ability to capture label dependencies. In this paper, we introduce a new class of loss functions for multi-label classification, which overcome disadvantages of commonly used losses such as Hamming and subset 0/1. To this end, we leverage the mathematical framework of non-additive measures and integrals. Roughly speaking, a non-additive measure allows for modeling the importance of correct predictions of label subsets (instead of single labels), and thereby their impact on the overall evaluation, in a flexible way---by giving full importance to single labels and the entire label set, respectively, Hamming and subset 0/1 are rather extreme in this regard. We present concrete instantiations of this class, which comprise Hamming and subset 0/1 as special cases, and which appear to be especially appealing from a modeling perspective. The assessment of multi-label classifiers in terms of these losses is illustrated in an empirical study.     

\end{abstract}
\section{Introduction}

The setting of multi-label classification (MLC), which generalizes standard multi-class classification by relaxing the assumption of mutual exclusiveness of classes, has received a lot of attention in the recent machine learning literature---we refer to \citet{DBLP:reference/dmkdh/TsoumakasKV10} and \citet{DBLP:journals/tkde/ZhangZ14} for survey articles on this topic. The motivation for MLC originated in the field of text categorization \cite{DBLP:conf/iaai/HayesW90,DBLP:conf/sigir/Lewis92,DBLP:journals/tois/ApteDW94}, but nowadays multi-label methods are used in applications as diverse as music categorization \cite{DBLP:journals/ejasmp/TrohidisTKV11}, semantic scene classification \cite{DBLP:journals/pr/BoutellLSB04}, and protein function classification \cite{DBLP:conf/pci/DiplarisTMV05}.

Formally, the task of a multi-label classifier is to assign a subset of a given set of candidate labels to any query instance. A straightforward approach for learning such a predictor is via a reduction to binary classification, i.e., by training one binary classifier per label and combining the predictions of these classifiers into an overall multi-label prediction. This approach, known as \emph{binary relevance} (BR) learning, is often criticized for ignoring possible label dependencies, because each label is predicted independently of all other labels. Indeed, the idea of exploiting statistical dependencies between the labels in order to improve predictive performance on the level of the entire label set is a major theme in research on multi-label classification, and many MLC methods proposed in the literature are motivated by this idea. 

Of course, the usefulness of such methods very much depends on whether or not there is a need to capture label dependence. This is not always the case, for example if such dependencies are indeed not present in the data. Besides, it turns out that the underlying loss function used to evaluate multi-label predictions plays an important role \cite{DBLP:journals/ml/DembczynskiWCH12}. Since a subset of predicted labels can be compared with a ground-truth subset in many ways, various loss functions have been proposed in the literature. Two simple but commonly used examples are the Hamming and the subset 0/1 loss, which are both generalizations of the 0/1 loss in conventional single-label classification. 
While the former assesses the quality of predictions as the percentage of incorrectly predicted labels, the latter measures the fraction of label subsets that are not predicted correctly in their entirety, i.e., for which at least one label is predicted incorrectly.
%While the former penalizes a prediction by the percentage of incorrectly predicted labels, the latter only checks whether or not the label subset is correct in its entirety, and correspondingly assigns a loss of 0 or 1. 
As will be explained in more detail later on, capturing label dependencies is crucial for performing well in terms of subset 0/1 loss, but---at least in theory---not necessary in the case of Hamming.%  (at least theoretically). 

\begin{comment} 
% I think this sentence is redundant. (JF)
Different loss functions are often motivated by the application at hand. Besides, they are used in a more systematic way to foster and to analyze a learner's ability to capture label dependencies. In particular, the subset 0/1 loss is commonly used for this purpose.  
\end{comment}
Despite their widespread use and interesting theoretical properties, both Hamming and subset 0/1 can be criticized for various reasons, especially for being rather extreme. The Hamming loss is often close to 0, simply because the label cardinality (average percentage of relevant labels per example) is very small in typical MLC data sets. Thus, even the default classifier that predicts all labels as irrelevant will usually perform well according to Hamming loss, and is indeed often difficult to beat. Even if an improvement is possible, the performance differences are typically small, and therefore difficult to test for statistical significance. On the other side, the subset 0/1 loss is normally quite high and may appear overly stringent, especially in the case of many labels. Moreover, since making a mistake on a single label is punished as hardly as a mistake on all labels, it does not discriminate well between ``almost correct'' and completely wrong predictions. 

To overcome disadvantages of commonly used losses such as Hamming and subset 0/1, we introduce a new class of loss functions for multi-label classification (Section~\ref{sec:non-additive-loss}). To this end, we leverage the mathematical framework of non-additive measures and integrals. Roughly speaking, the overall loss is obtained by integrating over the errors on individual labels. This integration is done with respect to a non-additive measure, which allows for controlling the ``dependence-awareness'' of the loss, i.e., for modeling the importance of correct predictions of label subsets. We present concrete instantiations of this type of loss functions, which 
%comprise Hamming and subset 0/1 as special cases, and which  appear to be especially appealing from a modeling perspective. 
allow for controlling dependence-awareness by means of a single parameter. The assessment of the ``dependence-awarenenss'' of multi-label classifiers in terms of these losses is illustrated in an empirical study (Section~\ref{sec:experiments}).

\section{Multi-label Classification}
\label{sec:multi-label}

Let $\cX$ denote an instance space, and let $\cL= \{\lambda_1, \ldots, \lambda_K\}$ be a finite set of class labels. We assume that an instance $\bx \in \cX$ is (probabilistically) associated with a subset of labels $\Lambda = \Lambda(\bx) \in 2^\cL$; this subset is often called the set of relevant labels, while the complement $\cL \setminus \Lambda$ is considered as irrelevant for $\bx$. We identify a set $\Lambda$ of relevant labels with a binary vector $\by = (y_1, \ldots, y_K)$, where $y_k = \llbracket \lambda_k \in \Lambda \rrbracket$.\footnote{$\llbracket \cdot \rrbracket$ is the indicator function, i.e., $\llbracket A \rrbracket = 1$ if the predicate $A$ is true and $=0$ otherwise.} By $\cY = \{0,1\}^K$ we denote the set of possible labelings.  

We assume observations to be realizations of random variables generated independently and identically (i.i.d.) according to a probability measure $\Prob$ on $\cX \times \cY$ (with density/mass function $\prob$), i.e., an observation $\by=(y_1,\ldots, y_K)$ is the realization of a corresponding random vector $\bY = (Y_1, \ldots, Y_K)$. We denote by $\prob(\mathbf{Y} \given \bx)$ the conditional distribution of $\bY$ given $\mathbf{X}=\bx$, and by $\prob_i(Y_i \given \bx)$ the corresponding marginal distribution of the $i$-th label $Y_i$:
\begin{equation}\label{eq:marginal}
\prob_i( b \given \bx) = %\Pr^{(i)}( b \, | \, \bx) = 
\sum_{\by\in\cY: y_i = b} \prob(\by \given \bx) \, .
 %(\by \, | \, \bx)
\end{equation}
Given training data in the form of a finite set of observations
\begin{equation}\label{eq:trainingdata}
\cD = \big\{ (\bx_n,\by_n) \big\}_{n=1}^N  \subset \cX \times \cY \, ,
\end{equation}
drawn independently from $\Prob(\mathbf{X},\mathbf{Y})$, the goal in MLC is to learn a predictive model that generalizes well beyond these observations, i.e., which yields predictions that minimize the expected risk with respect to a specific loss function. In this regard, we need to clarify what type of predictions are sought and how these predictions are assessed.

\subsection{Predictive Models in MLC}

A multi-label classifier $\bh$ is a mapping $\cX \fromto \cY$  that assigns a (predicted) label subset to each instance $\bx\in \cX$. Thus, the output of a classifier $\bh$ is a vector 
\begin{equation}\label{eq:h}
\bh(\bx) = (h_1(\bx), \ldots , h_K(\bx)) \in \{ 0,1 \}^K \,  .
\end{equation}
Predictions of this kind will also be denoted $\hat{\by} = (\hat{y}_1, \ldots , \hat{y}_K)$. 

Sometimes, MLC is treated as a \emph{ranking} (instead of a subset selection) problem, in which the labels are sorted according to their degree or probability of relevance. Then, the prediction takes the form of a  \emph{scoring function}:
\begin{equation}\label{eq:s}
\bs(\bx) = (s_1(\bx), s_2(\bx), \ldots , s_K(\bx)) \in \mathbb{R}^K \, .
\end{equation}
A prediction of that kind encodes a ranking $\pi:\, [K] \fromto [K]$, such that $\pi(i)$ is the position of label $\lambda_i$. This ranking is obtained by sorting the labels $\lambda_i$ in decreasing order according to their scores $s_i(\bx)$.  

%Often, multilabel classifiers are implemented in the form of thresholded scoring classifiers, which means that 
%$h_i(\bx) = \llbracket  f_i(\bx)) > t  \rrbracket$,
%where $f_i(\cdot)$ is a (real-valued) scoring function and $t$ a threshold.\footnote{For a predicate $P$, the expression $\llbracket P \rrbracket$ evaluates to 1 if $P$ is true and to 0 is $P$ is false. } The underlying scoring function
%\begin{equation}\label{eq:rf}
%\bf(\bx) = (f_1(\bx), f_2(\bx), \ldots , f_K(\bx)) 
%\end{equation}
%is also called \emph{ranking function}, as it can be used in an obvious way to predict a ranking of the labels. To this end, the labels $\lambda_i$ are simply sorted in decreasing order according to their scores $f_i(\bx)$. 

\subsection{MLC Loss Functions}

In the literature, various MLC loss functions have been proposed. Commonly used are the Hamming loss $\loss_H$ and the subset 0/1 loss $\loss_S$, which both generalize the standard 0/1 loss for multi-class classification, albeit in very different ways:
%\\
%\noindent\begin{tabularx}{\linewidth}{@{}XX@{}}
\begin{equation}
\label{eq:hamming}
\loss_H(\by, \hat{\by}) = \frac{1}{K}
\sum_{k=1}^K  \, \llbracket y_k \neq  \hat{y}_k \rrbracket \enspace 
\end{equation}
%&
%\medskip
\begin{equation}\label{eq:subset01}
\loss_S(\by, \hat{\by}) = \llbracket \by \neq  \hat{\by} \rrbracket \enspace 
\end{equation}
%\end{tabularx}
\begin{comment}
The Hamming loss is given by
\begin{equation}
\label{eq:hamming}
\loss_H(\by, \hat{\by}) = \frac{1}{K}
\sum_{k=1}^K  \, \llbracket y_k \neq  \hat{y}_k \rrbracket \enspace , 
\end{equation}
and the subset 0/1 loss by 
\begin{equation}\label{eq:subset01}
\loss_S(\by, \hat{\by}) = \llbracket \by \neq  \hat{\by} \rrbracket \enspace .
\end{equation}
\end{comment}
%Thus,  
Besides, other performance metrics are often reported in experimental studies. For example, the (instance-wise) F-measure is defined in terms of the harmonic mean of precision and recall, and can be written as follows:
$$
F(Y, \hat{Y})  = \frac{2 \sum_{k=1}^K \hat{y}_k \, y_k}{\sum_{k=1}^K \hat{y}_k + \sum_{k=1}^K y_k} 
$$
The F-measure takes values in the unit interval and can be turned into a loss function by setting $\loss_F(Y, \hat{Y}) = 1 - F(Y, \hat{Y})$. 

\subsection{Label Dependence}

The goal of classification algorithms in general is to capture dependencies between input features $X_i$ and the target variable $Y$. In fact, the prediction of a scoring classifier is often regarded as an approximation of  the conditional probability $\prob(Y = \hat{y} \given \bx)$, i.e., the probability that $\hat{y}$ is the true label for the given instance $\bx$. In MLC, dependencies may not only exist between the features $X_i$ and each target, but also between the targets $Y_1, \ldots , Y_K$ themselves. The idea to improve predictive accuracy by capturing such dependencies is a driving force in research on multi-label classification. 

In this regard, a distinction between \emph{unconditional} and \emph{conditional independence} of labels can be made \cite{DBLP:journals/ml/DembczynskiWCH12}. In the first case, the joint distribution $\prob(\bY)$ in the label space factorizes into the product of the marginals $\prob(Y_k)$, i.e., 
$$
\prob(\bY) = \prob(Y_1) \times \prob(Y_2) \times  \cdots \times \prob(Y_K) \, , 
$$
whereas in the latter case, the factorization 
$$
\prob(\bY \given \bx) = \prob(Y_1 \given \bx)  \times \prob(Y_2 \given \bx) \times \cdots \times \prob(Y_K \given \bx)
$$
holds conditioned on $\bx$, for every instance $\bx$.
In other words, unconditional dependence is a kind of global dependence (for example originating from a hierarchical structure on the labels), whereas conditional dependence is a dependence locally restricted to a single point in the instance space. 
%In addition to this probabilistic view on label dependence, some attempts at formalizing different types of rule-based dependencies have been made \cite{jf:PL-08-WS-Park,ML:Rules-Stacking}.

It turns out that there is a close connection between label dependence and the \emph{decomposability} of loss functions: A decomposable loss can be expressed in the form 
\begin{equation}\label{eq:decoml}
\loss(\by , \hat{\by} ) = \sum_{k=1}^K \loss_k(y_k , \hat{y_k} )
\end{equation}
with suitable binary loss functions $\loss_k:\, \{0,1\}^2 \fromto \mathbb{R}$, whereas a non-de\-com\-pos\-a\-ble loss does not permit such a representation. It can be shown that, to produce optimal predictions $\hat{\by} = \bh(\bx)$ minimizing expected loss, knowledge about the marginals $\prob_k(Y_k \given \bx)$ is enough in the case of a decomposable loss (such as Hamming), but not in the case of a non-decomposable loss \cite{DBLP:journals/ml/DembczynskiWCH12}. Instead, if a loss is non-decomposable, high-order probabilities are needed, and in the extreme case even the entire distribution $\prob(\mathbf{Y} \given \bx)$ (like in the case of the subset 0/1 loss). On an algorithmic level, this means that MLC with a decomposable loss can be tackled by binary relevance learning (i.e., learning one binary classifier for each label individually), whereas non-decomposable losses call for more sophisticated learning methods that are able to take label-dependencies into~account.

\section{MLC Loss Functions based on Non-Additive Measures}
\label{sec:non-additive-loss}

The Hamming and the subset 0/1 loss %are frequently used and commonly reported in experimental studies. Perhaps more importantly, they 
are often considered as prototypical examples of losses which, respectively, do and do not impel the learner to take label dependencies into account: Hamming is label-wise decomposable and can principally be optimized by learning algorithms like BR. The subset 0/1 loss, on the other side, is not label-wise decomposable. Therefore, this loss is often used to quantify the learner's ability to capture label dependencies. For example, consider the following (conditional) ground-truth distribution $\prob( \cdot \given \bx)$ on the label space $\cY = \{ 0,1 \}^3$:
%$$
%\Prob( \by \given \bx) = \left\{ \begin{array}{cl}
%\nicefrac{1}{4} & \text{if } \by = (1,1,1) \\
%\nicefrac{3}{16} & \text{if } \by \neq (1,1,1)
%\end{array} \right.
%$$
\begin{center}
\begin{tabular}{c|cccccccc}
$\by$ & $(0,0,0)$ & $(1,1,1)$ & $(0,1,1)$ & $(1,0,1)$ & $(1,1,0)$ \\
\hline
 $\prob( \by \given \bx)$ & $\nicefrac{1}{4}$ & $\nicefrac{3}{16}$ & $\nicefrac{3}{16}$ & $\nicefrac{3}{16}$ & $\nicefrac{3}{16}$
\end{tabular}
\end{center}
For each of the three labels, the individual probability of relevance is higher than the probability of irrelevance, and indeed, $\hat{\vec{y}} = (1,1,1)$ is the Bayes-optimal prediction (minimizing the loss in expectation) in the case of Hamming. For the subset 0/1 loss, however, the Bayes-optimal prediction is $\hat{\vec{y}} =(0,0,0)$. In general, the Bayes-optimal prediction is given by the \emph{marginal mode} of the distribution $\prob(\cdot \given \bx)$ in the case of Hamming and by the \emph{joint mode} in the case of subset 0/1.  

As already said, both Hamming and subset 0/1 can be criticized for being rather extreme. Due to the reasons already explained in the introduction (imbalance between relevant and irrelevant labels), the Hamming loss is often very low. As opposed to this, the subset 0/1 loss is normally quite high, since an entirely correct prediction becomes very unlikely with increasing $K$. It is an ``all or nothing'' measure, for which a mistake on a single label is as bad as a mistake on many labels, and which does not reward correct predictions on larger subsets of the labels. 
%and may indeed appear overly stringent, especially in the case of many labels. Moreover, since making a mistake on a single label is punished as hardly as a mistake on all labels, it does not discriminate well between ``almost correct'' and completely wrong predictions. 

%\begin{tabular}{cc}
%\hline
%$\by$ &     $\prob( \by \given \bx)$ \\
%\hline
%$(0,0,0)$ & $\nicefrac{1}{4}$ \\
%$(1,1,1)$ & $\nicefrac{3}{16}$ \\
%$(0,1,1)$ & $\nicefrac{3}{16}$ \\
%$(1,0,1)$ &  $\nicefrac{3}{16}$ \\
%$(1,1,0)$ & $\nicefrac{3}{16}$ \\
%\hline
%\end{tabular}

%\begin{tabular}{cccc}
%\hline
%$(y_i,y_j)$ &  $\prob( y_1, y_2 \given \bx)$ &  $\prob( y_1, y_3 \given \bx)$ &  $\prob( y_2, y_3 \given \bx)$\\
%\hline
%$(0,0)$ & $\nicefrac{1}{4}$ & $\nicefrac{1}{4}$  & $\nicefrac{1}{4}$  \\
%$(0,1)$ & $\nicefrac{3}{16}$ & $\nicefrac{3}{16}$ & $\nicefrac{3}{16}$ \\
%$(1,0)$ & $\nicefrac{3}{16}$ & $\nicefrac{3}{16}$ & $\nicefrac{3}{16}$ \\
%$(1,1)$ & $\nicefrac{6}{16}$ & $\nicefrac{3}{16}$ & $\nicefrac{3}{16}$ \\
%\hline
%\end{tabular}
%\begin{tabular}{cccc}
%\hline
%$y_1$ &     $\prob( y_1 \given \bx)$ &     $\prob( y_2 \given \bx)$ &     $\prob( y_3 \given \bx)$ \\
%\hline
%$0$ & $\nicefrac{7}{16}$ & $\nicefrac{7}{16}$ & $\nicefrac{7}{16}$ \\
%$1$ & $\nicefrac{9}{16}$ & $\nicefrac{9}{16}$ & $\nicefrac{9}{16}$ \\
%\hline
%\end{tabular}

To overcome these disadvantages, we introduce a new class of loss functions for multi-label classification in Section \ref{sec:gloss}. These loss functions are able to assess a learner's dependence-awareness, i.e., its aptness at capturing label dependencies, in a more skillful manner. To this end, we leverage the mathematical framework of non-additive measures and integrals, the essentials of which are recalled in Sections \ref{b10} and \ref{sec:ci}. Roughly speaking, a non-additive measure is used for modeling the importance of correct predictions of label subsets (instead of single labels), and thereby their impact on the overall evaluation. As will be seen, Hamming and subset 0/1 will be recovered as special cases of our family, which, in a sense, allows for ``interpolating'' between these two extremes. 
%Besides, we present other instantiations of this family, which appear to be especially appealing from a modeling perspective.

For didactic reasons, let us anticipate the basic construction principle of our family of loss functions, which will be introduced step by step alongside with a couple of other (auxiliary) functions. More specifically, considering the correctness of predictions on individual labels $\lambda_i$ as evaluation criteria $c_i$, a loss $\ell_\mu(\vec{y}, \vec{s})$ will be defined as a suitably weighted aggregation of the correctness degrees
\begin{equation}\label{eq:cofu}
f(c_i) =  1-|s_i - y_i| \in [0,1] \, ,
\end{equation}
where $s_i \in [0,1]$ is the score predicted for label $\lambda_i$ and $y_i \in \{ 0,1 \}$ the corresponding ground truth. Allowing for predictions in terms of a score vector $\vec{s} = (s_1, \ldots , s_K) \in [0,1]^K$ is more general than a binary prediction $\hat{\vec{y}} = (\hat{y}_1, \ldots , \hat{y}_K) \in \{0,1\}^k$, but obviously comprises the latter as a special case. The loss will then be specified in terms of an integral of the ``correctness function'' $f$ given by (\ref{eq:cofu}), i.e., as an aggregated (in-)correctness
\begin{equation}\label{eq:int}
\ell_\mu(\vec{y}, \vec{s}) = 1 -  \int f \, d \, \mu \, ,
\end{equation}
To this end, two main ingredients are needed, namely the measure $\mu$ for weighting and the integral for aggregation:
%one form modeling correctness of predictions, one for weighting the importance of label subsets, and one for aggregation. For didactic reasons, we provide a brief summary of these ingredients at this point:
\begin{itemize}
 %   \item The ``correctness function'' $f$ maps each label to the correctness of the prediction on that label. In general, since real-valued scores are allowed as predictions, correctness is a value in $[0,1]$. If predictions are binary, correctness is either 0 or 1.     
\item  A \emph{measure} $\mu$ assigns a weight $\mu(A)$ to every subset $A$, in our case to a subset of labels, which can be interpreted as the importance of that subset. Formally, a measure $\mu$ is a mapping from subsets to the unit interval, which can be equivalently represented by its \emph{M\"obius transform} $m_\mu$. 
\item The aggregation in (\ref{eq:int}) is accomplished with the so-called (discrete) \emph{Choquet integral} $\mathcal{C}_\mu$, which is a weighted aggregation of the values of a function (in our case $f$) with respect to the underlying measure $\mu$.  
%Thus, $\mathcal{C}_\mu(f)$ can be thought of as a weighted aggregation of the label-wise correctness of a prediction. 
%    \item Eventually, the \emph{loss function} $\ell_\mu$ is defined based on the Choquet-aggregation of the correctness function with respect to be measure $\mu$, namely as the complement of $\mathcal{C}_\mu(f)$. 
\end{itemize}
In the following, we discuss the components on the right-hand side of (\ref{eq:int}) in more detail.

\subsection{Non-Additive Measures}\label{b10}

Let $C=\{c_1, \ldots , c_K\}$ be a finite set of (desirable) ``criteria'' and $\mu: 2^C \longrightarrow [0,1]$ a measure on this set. For each $A \subseteq C$, we interpret $\mu(A)$ as the {\em weight} or, say, the \emph{importance} of the subset of criteria $A$. In the context of MLC, we can think of the criterion $c_i$ as the correctness of the prediction on the $i^{th}$ label $\lambda_i$. Thus, $\mu(\{c_1\})$ is the importance of predicting the first label correctly, and $\mu(\{c_1,c_2\})$ is the importance of \emph{jointly} predicting the first and the second label correctly.

%As an illustration that makes a link to machine learning, one may think of $X$ as a set of attributes in a data set, and of $\mu(A)$ as the (expected) performance of a specific classifier (e.g., a decision tree) when being trained on the subset $A$ of features. 
%As an illustration, one may think of $C$ as a set of criteria (binary features) relevant for a job, like ``speaking French'' and ``programming Java'', and of $\mu(A)$ as the evaluation of a candidate satisfying criteria $A$ (and not satisfying $C \setminus A$). The term ``criterion'' is indeed often used in the decision making literature, where it suggests a monotone ``the higher the better'' influence. In the context of machine learning, to which we shall turn later on, criteria are playing the role of features (input attributes).

A standard assumption on a measure $\mu$, which is at the core of probability theory, is additivity:
$\mu(A \cup B) = \mu(A) + \mu(B)$
for all $A,B \subseteq C$ such that $A \cap B = \emptyset$.
Unfortunately, additive measures cannot model any kind of ``interaction'': Extending a set of elements $A$
by a set of elements $B$ always increases
the weight $\mu(A)$ by the weight $\mu(B)$, regardless of $A$ and $B$. For example, we cannot express that predicting $\lambda_1$ and $\lambda_2$ correctly, i.e., both together, has a higher value than the sum of getting both of them individually right.

%Suppose, for example, that the elements of two sets $A$ and $B$ are {\em complementary} in a certain sense. For instance, $A = \{ \mathtt{French}, \mathtt{Spanish}\}$ and $B = \{ \mathtt{Java} \}$ could be seen as complementary, since both language skills and programming skills are important for the job.  Formally, this can be expressed in terms of a positive interaction: $\mu(A \cup B) > \mu(A) + \mu(B)$. In the extreme case, when language skills and programming skills are indeed essential, $\mu(A \cup B)$ can be high although $\mu(A) = \mu(B) = 0$ (suggesting that a candidate lacking either language or programming skills is completely unacceptable). Likewise, elements can interact in a negative way: If two sets $A$ and $B$ are partly {\em redundant} or {\em competitive}, then $\mu(A \cup B) < \mu(A) + \mu(B)$. For example,  $A = \{ \mathtt{C}, \mathtt{C}\# \}$ and $B = \{ \mathtt{Java} \}$ might be seen as redundant, since one programming language does in principle suffice.   

%The above considerations motivate the use of 
Non-additive measures,
also called capacities or fuzzy measures,
are simply normalized and monotone, but not necessarily additive \cite{suge_to}:
%Such measures,
%which are often called {\em fuzzy measures} in the literature \cite{suge_to}, satisfy
%the following properties:
\begin{equation}\label{eq:nam}
\begin{split}
&\mu(\emptyset) = 0, \mu(C) = 1, \text{ and}\\
&\mu(A) \leq \mu(B) \text{ for all } A \subseteq B \subseteq C \; .
\end{split}
\end{equation}
Thus, a set of criteria $B$ is always at least as important as any of its subsets. 

%The first property guarantees a fuzzy measure to be normalized.\footnote{The
%condition $\mu(X)=1$ is often not necessary and sometimes omitted.}
%The second property expresses the monotonicity of a fuzzy measure:
%The weight of a set $A$ can never decrease by adding further elements.
%
A useful representation of non-additive measures is in terms of the \emph{M\"obius transform}:
\begin{equation}\label{eq:mt1}
\mu(B) = \sum\limits_{A\subseteq{B}} \moebius_{\mu}(A)
\end{equation}
for all $B \subseteq C$, where the M\"obius transform $\moebius_{\mu}$ of the measure $\mu$ is defined as follows:
\begin{equation}\label{eq:mt}
     \moebius_{\mu}(A) = \sum\limits_{B\subseteq{A}}(-1)^{|A|-|B|}\mu{(B)} \; .
\end{equation}
The value $\moebius_{\mu}(A)$ can be interpreted as the weight that is \emph{exclusively} allocated to $A$, instead of being indirectly connected with $A$ through the interaction with other subsets.
 
A measure $\mu$ is said to be $k$-order additive, or simply $k$-additive, if $k$ is the smallest integer such that $\moebius(A)=0$ for all $A \subseteq C$ with $|A| > k$. This property is interesting for several reasons. First, as can be seen from (\ref{eq:mt1}), it means that a measure $\mu$ can formally be specified by 
%at most 
%\[
%\sum_{i=1}^k \left(\begin{array}{cc} 
%m \\ i 
%\end{array} \right)  
%\]
significantly fewer than $2^K$ values, which are needed in the general case. Second, $k$-additivity is also interesting from a semantic point of view, as it means that there are no interaction effects between subsets $A,B \subseteq C$ whose cardinality exceeds $k$.

\subsection{The Choquet Integral}
\label{sec:ci}

%So far, the criteria $c_i$ were simply considered as binary features, which are either present or absent. Mathematically, $\mu(A)$ can thus also be seen as an \emph{integral} of the indicator function of $A$, namely the function $f_A$ given by $f_A(c)=1$ if $c \in A$ and $=0$ otherwise. 
%Now, 
Suppose that $f:\, C \longrightarrow \mathbb{R}_+$ is a non-negative function that assigns a \emph{value} to each criterion $c_i$. In the case of MLC, we can think of $f(c_i)$ as the correctness of a prediction on the label $\lambda_i$. 
%for example, $f(c_i)$ might be the degree to which a candidate satisfies criterion $c_i$. 
An important question, then, is how to \emph{aggregate} the evaluations of individual criteria, i.e., the values $f(c_i)$, into an overall evaluation, in which the criteria are properly weighted according to the measure $\mu$. Mathematically, this overall evaluation can be considered as an integral $\mathcal{C}_\mu(f)$ of the function $f$ with respect to the measure $\mu$.

Indeed, if $\mu$ is an additive measure, the standard integral just corresponds to the \emph{weighted mean}
\begin{equation}\label{eq:wm}
\mathcal{C}_\mu(f) = \sum_{i=1}^K w_i \cdot f(c_i)  = \sum_{i=1}^K \mu(\{c_i\}) \cdot f(c_i)
\; , 
\end{equation}
which is a natural aggregation operator in this case. For example, in the context of MLC, the Hamming loss is a special case of (\ref{eq:wm}), with $f(c_i) \in \{0,1\}$ depending on whether the prediction on $\lambda_i$ is right or wrong, and uniform weights $w_i = 1/K$.  

\begin{figure}[tb]
	\centering 
	\includegraphics[width=0.75\columnwidth]{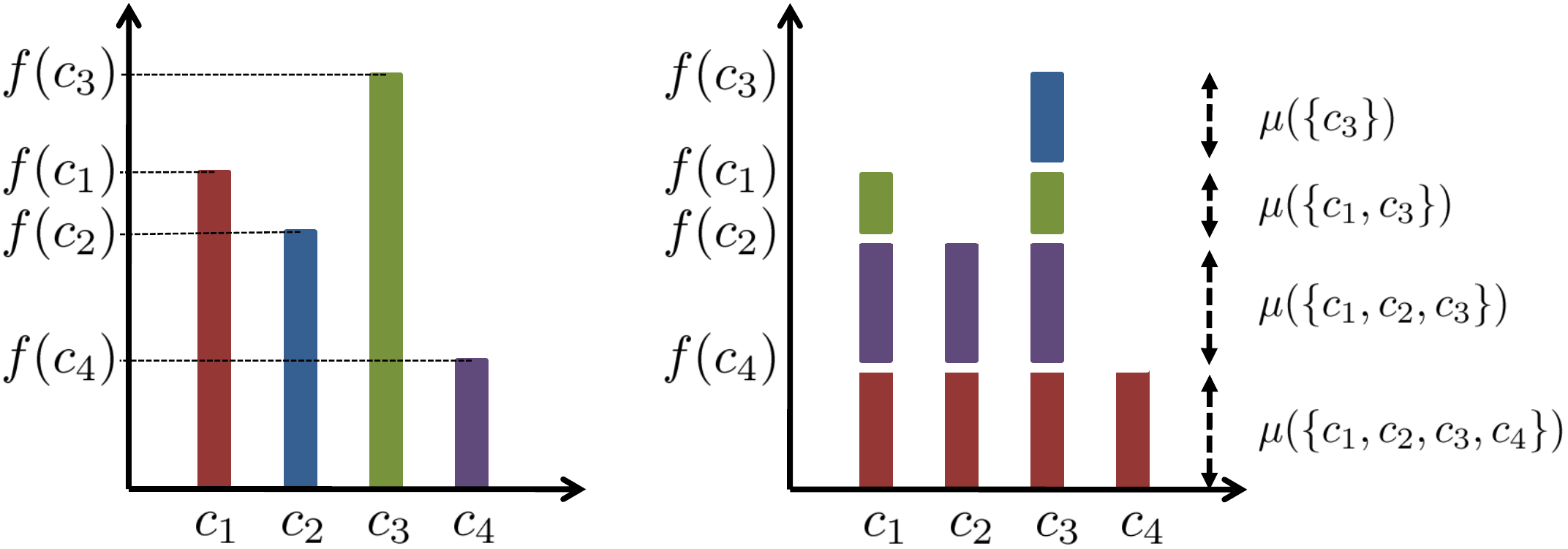}
	\caption{Vertical (left) versus horizontal (right) integration. In the first case, the height of a single bar, $f(c_i)$, is multiplied with its ``width'' (the weight $\mu(\{c_i\})$, and these products are added. In the second case, the height of each horizontal section, $f(c_{(i)})-f(c_{(i-1)})$, is multiplied with the corresponding ``width'' $\mu(A_{(i)})$.} 
	\label{fig1}
\end{figure}

A non-trivial question, however, is how to generalize (\ref{eq:wm}) in the case where $\mu$ is non-additive. This question, namely how to define the integral of a function with respect to a non-additive measure (not necessarily restricted to the discrete case), is answered in a satisfactory way by the Choquet integral \cite{choquet1954theory}. The point of departure of the Choquet integral is an alternative representation of the ``area'' under the function $f$, which, in the additive case, is a natural interpretation of the integral. Roughly speaking, this representation decomposes the area in a ``horizontal'' instead of a ``vertical'' manner, thereby making it amenable to a straightforward extension to the non-additive case. More specifically, note that the weighted mean can be expressed as follows:
\begin{align*}
\sum_{i=1}^K f(c_i) \cdot \mu(\{c_i\})  = & \sum_{i=1}^K \Big( f(c_{(i)})-f(c_{(i-1)}) \Big) \Big( \mu(\{c_{(i)}\}) + \ldots + \mu(\{c_{(K)}\}) \Big)\\
%& = \sum_{i=1}^K (f(x_{(i)})-f(x_{(i-1)}) \cdot \mu(\{ x \, | \,  f(x) \geq f(x_{(i)}) \}))\\
= & \sum_{i=1}^K \Big( f(c_{(i)})-f(c_{(i-1)}) \Big) \cdot \mu \Big(A_{(i)} \Big) \; ,
\end{align*}
where $(\cdot)$ is a permutation of $[K]$ such that 
$0 \le f(c_{(1)}) \le f(c_{(2)}) \le \ldots \le f(c_{(K)})$ (and $f(c_{(0)})=0$ by definition), and $A_{(i)}= \{c_{(i)}, \ldots , c_{(K)}\}$; see Fig.\ \ref{fig1} for an illustration. 

Now, the key difference between the left and right-hand side of the above expression is that, whereas the measure $\mu$ is only evaluated on single elements $c_i$ on the left, it is evaluated on \emph{subsets} of elements on the right. 
%In fact, the right-hand side can formally be computed for a non-additive measure, too, and doing so will 
Thus, the right-hand side suggests an immediate extension to the case of non-additive measures, namely the Choquet integral, which, in the discrete case, is formally defined as follows:
\[
\mathcal{C}_\mu(f) = \sum\limits_{i=1}^K \left( f(c_{(i)})-f(c_{(i-1)}) \right) \cdot \mu(A_{(i)}) \ 
\]
\enlargethispage*{12pt}
A simple derivation shows that, in terms of the M\"obius transform of $\mu$, the Choquet integral can also be expressed as % follows:
\begin{align}
  \mathcal{C}_\mu(f)= 
  %& \sum\limits_{i=1}^K  \left( f(c_{(i)})-f(c_{(i-1)}) \right) \cdot \mu(A_{(i)}) \nonumber \\
  %= & \sum\limits_{i=1}^K  f(c_{(i)}) \cdot (\mu(A_{(i)})-\mu(A_{(i+1)}))  \nonumber \\
  %= & \sum\limits_{i=1}^K  f(c_{(i)}) \sum\limits_{R \subseteq T_{(i)}} \moebius(R)  \nonumber \\
   \sum\limits_{T\subseteq C} \moebius(T) \times \min_{i \in{T}}{f(c_{i})} \, .
   \label{eq:iam}
\end{align}
%where $T_{(i)} = \big\lbrace S \cup \{c_{(i)}\} \, | \, S \subset \{c_{(i+1)}, \ldots , c_{(m)} \} \big\rbrace$.

\subsection{MLC Loss Functions based on Non-Additive Measures}
\label{sec:gloss}

In the context of MLC, non-additive measures and generalized integrals can be used to define flexible loss functions: Each criterion $c_i$ corresponds to the (correct) prediction on a label $\lambda_i$, and $\mu(A)$ quantifies the importance to be correct on the subset of labels $A$ \emph{as a whole}. Moreover, the  function to be integrated is the correctness function (\ref{eq:cofu}).
Thus, $u_i = f(c_i) =  1-|s_i - y_i|$ is the degree of correctness on the label $\lambda_i$, where $s_i \in [0,1]$ is the score predicted for label $\lambda_i$ and $y_i \in \{ 0,1 \}$ the corresponding ground truth: $u_i= 1$ for a perfectly correct prediction and $u_i= 0$ for a completely wrong prediction. 
%As an aside, note that a prediction in terms of a score vector $\vec{s} = (s_1, \ldots , s_K) \in [0,1]^K$ is more general than a binary prediction $\hat{\vec{y}} = (\hat{y}_1, \ldots , \hat{y}_K) \in \{0,1\}^k$, but obviously comprises the latter as a special case.

Now, given the $u_i$ as values on the criteria $c_i$ (the higher the better), the idea is to aggregate these values with the Choquet integral (based on the measure $\mu$) into an overall degree of correctness, and to define a loss as the complement ($1-(\cdot)$) of this degree of correctness. Formally, this leads to 
%For a given measure $\mu$, we then obtain the performance metric 
\begin{equation}\label{eq:perf}
\loss_\mu(\vec{y} , \vec{s}) = 1 - \sum\limits_{i=1}^K \left( u_{(i)}- u_{(i-1)} \right) \cdot \mu(A_{(i)}) \ ,
\end{equation}
where the permutation $(\cdot)$ is such that 
$0 \le u_{(1)} \le u_{(2)} \le \ldots \le u_{(K)}$, and $A_{(i)}= \{c_{(i)}, \ldots , c_{(K)}\}$. %From (\ref{eq:perf}), we can derive a loss function that is parametrized by the measure $\mu$:
%\begin{equation}\label{eq:muloss}
%\loss_\mu(\vec{y} , \vec{s}) = 1 - g_\mu(\vec{y} , \vec{s}) \enspace .
%\end{equation}

\medskip
\noindent
\textbf{Special cases.}
Important special cases include the additive measure $\mu(A) = |A|/K$, for which we obtain
$$
\loss_\mu(\vec{y} , \vec{s}) = 1 - \frac{1}{K} \sum\limits_{i=1}^K  u_{i} = 
\frac{1}{K} \sum\limits_{i=1}^K  |s_i - y_i| \, ,
$$
i.e., the Hamming loss (or, strictly speaking, a generalization of the Hamming loss in the case of real-valued scores $s_i \in [0,1]$), and the measure $\mu$ defined by $\mu(C)=1$ and $\mu(A) = 0$ for $A \subsetneq C$, for which we obtain
$$
\loss_\mu(\vec{y} , \vec{s}) = \min_{1 \leq i \leq K}  u_{i} = 
\max_{1 \leq i \leq K}  |y_i - s_i  | \, ,
$$
i.e., the subset 0/1 loss (or again a generalization). 

%As another example, consider the measure $\mu$ defined by $\mu(A) = 1$ if $|A| \geq k$ and $\mu(A) = 0$ otherwise. Thus, $\mu$ gives full importance to $k$-subsets, but does not reward correctness on label subsets of size smaller $k$. In this case, we obtain 

Another interesting special case is the \emph{covering error}  introduced by \citet{amit_ab07}.
The latter is defined as the sum of subset 0/1 losses on a family of predefined label subsets, called a covering. The connection to this loss can nicely be seen based on the representation (\ref{eq:iam}) of the Choquet integral in terms of the M\"obius transform. Here, the $\min$-terms correspond to subset 0/1 losses on subsets $T$. In contrast to the covering error, where these losses are weighted equally, they are weighted by the values of the M\"obius function in our case.

\medskip
\noindent
\textbf{Counting measures.}
The two measures above are examples of so-called \emph{counting measures}, which only depend on the cardinality of $A$. In other words, $\mu$ is a counting measure if it can be expressed as $\mu(A) = v(\nicefrac{|A|}{K})$ for a suitable function $v:\, [0,1] \longrightarrow [0,1]$, which means that the measure of a set only depends on its cardinality but not the elements of the set. For example, $\mu(\{c_1,c_2\}) = v(2/K) =  \mu(\{c_3,c_4\})$. This kind of symmetry property is certainly meaningful in MLC, where the different labels are normally considered as equally important\,---\,or, stated differently, the performance metric is normally invariant under permutation of the labels. Here, $v(k/K)$ can be interpreted as the importance of a correct prediction on a subset of $k$ labels, which means that the loss function (\ref{eq:perf}) is completely specified by the values $0 = v(0), v(\nicefrac{1}{K}), \ldots , v(1)=1$.

Formally, for an increasing function $v:\, [0,1] \longrightarrow [0,1]$ such that $v(0)=0$ and $v(1)=1$, we obtain an OWA (ordered weighted averaging) \cite{DBLP:journals/tsmc/YagerF99,yager2012ordered} aggregation of the degrees of correctness $u_i$, namely
\begin{equation}\label{eq:owaloss}
 \sum_{i=1}^{K} w_i \cdot u_{(i)}  
\end{equation}
with 
$$
w_i =  v \left(\frac{K-i+1}{K} \right) - v \left(\frac{K-i}{K} \right)    \, .
$$
In other words,  we obtain an OWA loss function
\begin{equation}
\loss_\mu(\vec{y} , \vec{s}) = \sum_{i=1}^{K} w_i \cdot | y_{(i)} -  s_{(i)} | 
\end{equation}
with $w_1 + \ldots + w_K = 1$.
Again, Hamming is obtained for the special case $v:\, x \mapsto x$ and subset 0/1 for $v$ such that $v(x)=1$ for $x=1$ and $v(x)=0$ otherwise. 
Let us highlight that, in spite of a somewhat involved derivation (based on non-additive measures and integrals) and the flexibility our class of loss functions in general, the form (\ref{eq:owaloss}) we end up with in the case counting measures is both intuitively appealing and easy to compute. In principle, it is nothing than a weighted average of the errors on individual labels, with the important difference that the weights $w_i$ now pertain, not to the $i^{th}$ label, but to the $i^{th}$ order statistic of the error, i.e., the $i^{th}$ largest error. Let us illustrate this with a simple example, in which the ground-truth labeling is $\vec{y} = (0,1,1,0,0,0)$ and the prediction $\vec{s} = (0.2,0.3,0.9,0.1,0.4,0.3)$. Here, the errors on the individual labels are given, respectively, by $0.2, 0.7, 0.1, 0.1,  0.4, 0.3$. Sorting these from lowest to highest yields the increasing sequence $0.1, 0.1, 0.2, 0.3, 0.4, 0.7$. Different weight vectors $\vec{w}$ will then emphasize different values in this sequence and hence yield different losses, for example:

\bigskip

\begin{center}
\begin{tabular}{lccccccc}
\toprule
error  & 0.1 & 0.1 & 0.2 & 0.3 & 0.4 & 0.7  & $\loss_\mu(\vec{y} , \vec{s})$\\
\midrule
weight & $\nicefrac{1}{6}$ & $\nicefrac{1}{6}$ &$\nicefrac{1}{6}$ &$\nicefrac{1}{6}$ &$\nicefrac{1}{6}$ &$\nicefrac{1}{6}$ & 0.30 \\
weight & 0 & 0 &0 &0&0 &  1 & 0.70 \\
weight & $0$ & $\nicefrac{1}{15}$ &$\nicefrac{2}{15}$ &$\nicefrac{3}{15}$ &$\nicefrac{4}{15}$ &$\nicefrac{5}{15}$ & 0.43\\
\bottomrule
\end{tabular}
\end{center}

\bigskip

The first case with uniform weights corresponds to Hamming loss and yields a simple averaging of the errors. In the second case, the full weight is given to the largest error, which corresponds to the subset 0/1 loss. The third case is in-between these two extremes.

Let us also note that the computation is further simplified in the case of binary predictions, i.e., where the scores $s_i$ and hence also the individual errors are either 0 or 1. In this case, the loss merely depends on the total number of errors $k$, and  is given by
$$
\loss_\mu(\vec{y} , \vec{s}) = \sum_{i=K-k+1}^{K} w_{i} \, , 
$$
i.e., by the sum of the $k$ largest weights. 

\subsection{Parameterized Families}
In the following, we present two families of such loss functions, which allow for modeling the dependence-awareness in terms of a single parameter.  
\begin{itemize}
\item
\textbf{Polynomial loss}: First, one could think of using a convex function of the form 
\begin{equation}\label{eq:family1}
v:\, x \mapsto x^\alpha 
\end{equation}
for $\alpha \geq 1$. The larger $\alpha$, the more important it becomes to predict larger subsets correctly, and subset 0/1 is recovered for the limit case $\alpha \rightarrow \infty$. In other words, $\alpha$ can be used to smoothly interpolate between Hamming and subset 0/1. 
\item
\textbf{Binomial loss}: To motivate a second family of losses, suppose we are only interested in getting $k$-subsets of labels right, whereas a correct prediction on a subset of size $< k$ should not be rewarded. This could be reflected by a M\"obius function of the form 
$$
m(A) = \left\{
\begin{array}{cl}
 1/\binom{K}{k} & \text{ if } |A| = k\\[3mm] %\left( \begin{matrix} K \\ k \end{matrix} \right)^{-1} 
0 & \text{ otherwise }
\end{array} \right.
$$
In this case, we obtain 
\begin{equation}\label{eq:family2}
v\left( \frac{j}{K} \right) = \dfrac{\binom{j}{k}}{\binom{K}{k}} \, .
%\left( \begin{matrix} j \\ k \end{matrix} \right)  \left( \begin{matrix} K \\ k \end{matrix} \right)^{-1} \, .
\end{equation}
Again, the Hamming and subset 0/1 loss can be recovered by setting, respectively, $k=1$ and $k=K$, while interpolations are obtained in-between. % (see Section \ref{sec:kexample} for an example).

In principle, non-symmetric measures could of course be used in MLC as well, for example to express that different labels or different label subsets are of different importance. Yet, as already said, symmetry appears to be a natural property. Moreover, as it significantly reduces the number of degrees of freedom, this property facilitates the specification of a measure-based loss function (\ref{eq:perf}). 

What could nevertheless be interesting is a weighting of label subsets in proportion to the number of relevant labels they contain. More concretely, starting from a ``base measure'' $\mu$, the M\"obius mass  $m_\mu(A)$ could be adjusted depending on the number of relevant labels in $A$\,---\,increased if $A$ contains many and reduced if it contains only few relevant labels. Thereby, more emphasis could be put on correct predictions for relevant labels. The resulting loss function would then depend on the ground truth $\vec{y}$.

\end{itemize}

As 
% Here, we present 
an example of loss minimization for the Binomial loss, i.e., the loss (17) with $v$ given by (19), consider the following  
%Suppose the 
distribution on labelings $\vec{y}$ (given an instance $\vec{x}$):
% is as follows:

\begin{small}
\begin{center}
\begin{tabular}{cccccl}
\hline 
$y_1$ & $y_2$ & $y_3$ & $y_4$ & $y_5$ & $p(\vec{y} \given \vec{x})$ \\
\hline  
   0  &   0  &   0   &  0  &   0   &  0.046\\
   0  &   0  &   0   &  0  &   1   &  0.003\\
   0  &   0  &   0   &  1  &   0   &  0.034\\
   0  &   0  &   0   &  1  &   1   &  0.048\\
   0  &   0  &   1   &  0  &   0   &  0.025\\
   0  &   0  &   1   &  0  &   1   &  0.052\\
   0  &   0  &   1   &  1  &   0   &  0.036\\
   0  &   0  &   1   &  1  &   1   &  0.050\\
   0  &   1  &   0   &  0  &   0   &  0.022\\
   0  &   1  &   0   &  0  &   1   &  0.011\\
      0  &   1  &   0   &  1  &   0   &  0.006\\
   \hline
\end{tabular} \hfill
\begin{tabular}{cccccl}
\hline 
$y_1$ & $y_2$ & $y_3$ & $y_4$ & $y_5$ & $p(\vec{y} \given \vec{x})$ \\
\hline  
   0  &   1  &   0   &  1  &   1   &  0.059\\
   0  &   1  &   1   &  0  &   0   &  0.041\\
   0  &   1  &   1   &  0  &   1   &  0.023\\
   0  &   1  &   1   &  1  &   0   &  0.013\\
   0  &   1  &   1   &  1  &   1   &  0.012\\
   1  &   0  &   0   &  0  &   0   &  0.044\\
   1  &   0  &   0   &  0  &   1   &  0.023\\
   1  &   0  &   0   &  1  &   0   &  0.018\\
   1  &   0  &   0   &  1  &   1   &  0.011\\
1  &   0  &   1   &  0  &   0   &  0.003\\
   1  &   0  &   1   &  0  &   1   &  0.022\\
   \hline
\end{tabular} \hfill
\begin{tabular}{cccccl}
\hline 
$y_1$ & $y_2$ & $y_3$ & $y_4$ & $y_5$ & $p(\vec{y} \given \vec{x})$ \\
\hline  
   1  &   0  &   1   &  1  &   0   &  0.062\\
   1  &   0  &   1   &  1  &   1   &  0.054\\
   1  &   1  &   0   &  0  &   0   &  0.056\\
   1  &   1  &   0   &  0  &   1   &  0.059\\
   1  &   1  &   0   &  1  &   0   &  0.040\\
   1  &   1  &   0   &  1  &   1   &  0.022\\
   1  &   1  &   1   &  0  &   0   &  0.029\\
   1  &   1  &   1   &  0  &   1   &  0.013\\
   1  &   1  &   1   &  1  &   0   &  0.038\\
   1  &   1  &   1   &  1  &   1   &  0.025\\
   \\
   \hline
\end{tabular}
\end{center}
\end{small}

One can then verify (e.g., through simple enumeration) that the Bayes-optimal predictions for the Binomial loss with different parameters $k$ are given as follows:

\begin{center}
\begin{tabular}{lccccc}
   \hline
k=1:  & 1 &  0 &  0  & 1  & 0 \\
k=2:  & 0 &  0 &  1  & 1  & 1 \\
k=3:  & 0 &  0 &  1  & 1  & 1 \\
k=4:  & 1 &  1 &  0  & 0  & 0 \\
k=5:  & 1 &  0 &  1  & 1  & 0 \\
   \hline
\end{tabular}
\end{center}
This example shows that, by changing the parameter of the loss, the optimal prediction may change quite drastically. For example, the prediction of three of the five labels changes when going from $k=1$ to $k=2$, and even all five labels change when passing from $k=3$ to $k=4$.

\section{Empirical Case Study}
\label{sec:experiments}
In this section, we showcase how the proposed class of multi-label loss functions can be applied as an analysis tool for capturing the ``dependence-awareness'' of different multi-label classifiers, i.e., for assessing a learner's ability to capture label dependence. 

\subsection{Experimental Setup}
For the comparison of different multi-label classifiers, we apply them to various benchmark datasets originating from different domains.
In Table~\ref{tab:dataset}, an overview of the considered datasets together with their statistical properties is provided.
This includes the number of instances, the number of labels, the ratio of number of labels to number of instances, the absolute number of unique label combinations, and the average number of relevant labels per instance, also referred to as label cardinality.

\begin{table}[t!]
    \centering
    \caption{Overview of datasets with statistics of their main properties.}
    % to describe the properties of the datasets used in this empirical case study.}
    \label{tab:dataset}
    \begin{tabular}{p{1.5cm} l p{1.4cm} p{2.8cm} p{2.3cm} l}
    \toprule
        Dataset&\#Instances&\#Labels&Label-to-Instance\newline Ratio&Unique Label\newline Combinations& Cardinality\\
        \midrule
        birds & 645 & 19 & 0.0295 & 133 & 1.01\\
        emotions & 593 & 6 & 0.0101 & 27 & 1.87\\
        enron-f & 1702 & 53 & 0.0311 & 753 & 3.38\\
        flags & 194 & 12 & 0.0619 & 103 & 4.12\\
        genbase & 662 & 27 & 0.0408 & 32 & 1.25\\
        llog-f & 1460 & 75 & 0.0514 & 304 & 1.18\\
        medical & 978 & 45 & 0.0460 & 94 & 1.25\\
        scene & 2407 & 6 & 0.0025 & 15 & 1.07\\
        yeast & 2417 & 14 & 0.0058 & 198 & 4.24\\
        \bottomrule
    \end{tabular}
\end{table}

%Experiments are performed on nodes equipped with two Intel Xeon Gold 6148 and 192GB RAM and done in two steps.
We use paired 10-fold cross-validations for obtaining out-of-sample
%First, we split the datasets via 10-fold cross validation into training and test data, ensuring that all the classifiers are evaluated on the same splits.
%Second, we train each classifier for the respective training folds and obtain 
predictions in the form of label relevance scores.
% for the test data.
%The predicted label relevance score vectors are then merged for each method and considered as a whole for every dataset.
%We furthermore threshold each label at $\tau=0.5$.
%Hence, we subsequently compute the loss only considering scores $s_i \in \{0,1\}$.
Although
% In this experimental study, 
we restrict our analysis to binary predictions $s_i \in \{0,1\}$ in order to isolate from the ability of the classifiers to shape their scores, 
% In principle, however, 
our methodology is 
in principle 
also suitable for comparing soft predictions $s_i \in [0,1]$ and % therefore being 
independent of the thresholding technique used.

\subsection{Methods}

We experiment with several publicly available multi-label algorithms:
\begin{itemize}
\item 
\emph{Binary Relevance (BR)}  is a reduction to binary classification, which learns one binary classifier for each label independently of the others \cite{DBLP:journals/pr/BoutellLSB04}. Despite its simplicity, BR has proven to be highly competitive in comparison to state-of-the-art multi-label learners in recent studies, especially regarding measures that are not dependence-aware (cf.~e.g.~\cite{rivolli2020empirical,libre,wever2018automated}).
\item
\emph{Classifier Chains (CC)}  take label dependencies into account, by imposing an order on the label set and using the predictions for the previous labels as additional feature information for the next label predictor \cite{DBLP:conf/pkdd/ReadPHF09}.
\item
\emph{Label Powerset (LP)} is a reduction to multi-class classification \citep{DBLP:reference/dmkdh/TsoumakasKV10}. It
converts each possible label subset into a separate (meta-)class and then solves a standard classification problem. Thereby, it takes label dependence into account, though at the expense of treating similar label sets as independent classes. 
\item
\emph{Random k-Labelsets (RAkEL)}  
randomly selects several label subsets of a given size $k$, learns a (LP) multi-label classifier for each subset, and combines their predictions \cite{DBLP:conf/ecml/TsoumakasV07}. This may be viewed as a generalization of binary relevance ($K$ classifiers with $k=1$) and label powerset (1 classifier with $k=K$). Obviously, the larger $k$, the more dependence-aware this method should be.
\item
\emph{Predictive Clustering Trees (PCT)} 
build up a multi-objective decision tree by using example variance and multi-label prediction quality for guiding the tree construction \cite{DBLP:conf/ecml/KocevVSD07}. Full label vectors are predicted at the leafs, hence PCT allows a certain control over the dependence-awareness by setting the leaf and ensemble sizes.
\end{itemize}
%For PCTs, we used the implementation provided by Mulan\footnote{\url{https://mulan.sourceforge.net/}}, whereas the remaining implementations are taken from MEKA \cite{meka}. Both libraries extend the machine learning workbench WEKA \cite{DBLP:books/sp/datamining2005/FrankHHKP05}.
For all algorithms we used the implementations of MEKA, except PCTs, for which we used the implementation in  Mulan\footnote{\url{https://mulan.sourceforge.net/} and \url{http://waikato.github.io/meka/}.}.
Due to their favorable runtime, we used decision trees as single-label base learners in all MEKA methods.
Except for RAkEL, which is evaluated for different values $k$ and the number of ensemble members $m$, and PCT, which is used with single trees (PCT) and bagged ensembles of 10 trees (EPCT), all hyper-parameters are set to their default values.

\begin{figure}[t]
\begin{center}
\begin{subfigure}[c]{0.32\textwidth}
\centering $\ell_{bin}$
\includegraphics[width=\textwidth,trim={1cm 0.1cm 0 0},clip,page=1]{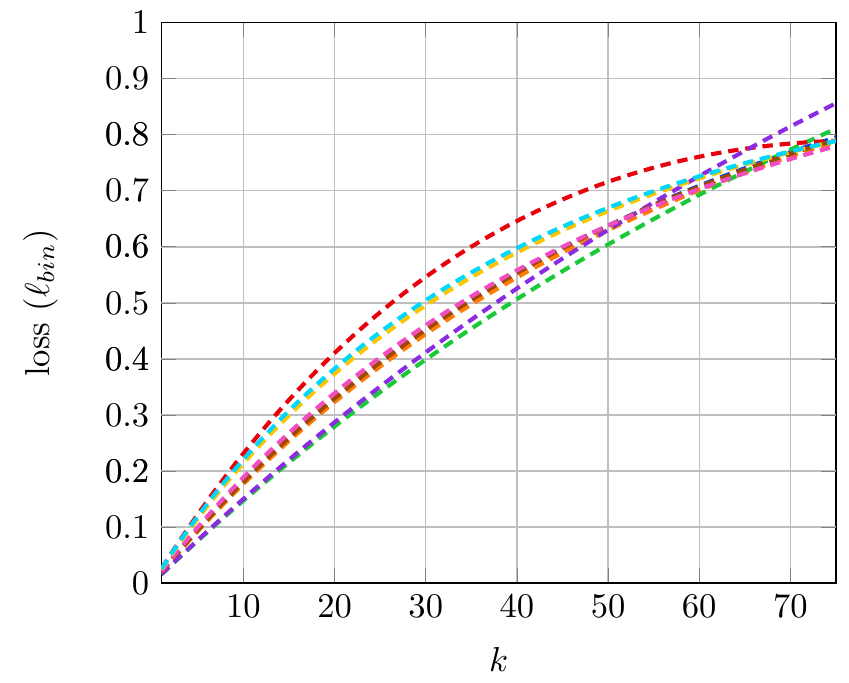}
\end{subfigure}
\hspace{0.5cm}
\begin{subfigure}[c]{0.32\textwidth}
\centering $\ell_{poly}$
\includegraphics[width=\textwidth,trim={1cm 0 0.1cm 0},clip,page=10]{datasetWisePlots.pdf}
\end{subfigure}

\includegraphics[width=.8\textwidth]{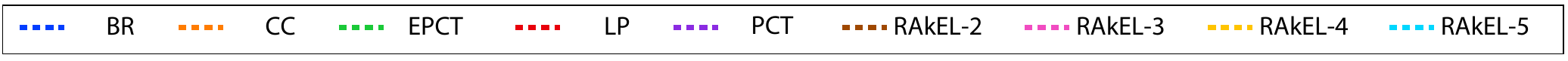}
\end{center}
\caption{Comparing multi-label algorithms on the dataset llog-f w.r.t.\ the binomial loss (left) and polynomial loss (right). 
}
\label{fig:datasetwise}
\end{figure}

\subsection{Results}
In the following, we present a selection of the results we produced and highlight several interesting insights. For a more comprehensive and detailed presentation, we refer to the supplementary material.

To analyze the dependence-awareness of the considered multi-label algorithms, we evaluate their performance in terms of the polynomial instantiation (\ref{eq:family1}) of our loss function, as well as the binomial instantiation (\ref{eq:family2})\,---\,we denote the former by $\ell_{poly}$ and the latter by $\ell_{bin}$.
While the (discrete) parameter $k$ of $\ell_{bin}$ takes values in $\{ 1 , \ldots , K\}$, we vary the (continuous) parameter of the polynomial loss, $\alpha$, between $1$ and $1000$.
In both cases, the lowest parameter value 1 corresponds to the Hamming loss and the highest values to the subset 0/1 loss (in the case of $\ell_{poly}$, strictly speaking, only for $\alpha \rightarrow \infty$), whereas intermediate values interpolate between these two extremes.
%Note that in the case of the polynomial loss family, a value of $1000$ for $\alpha$ does not necessarily yield the subset 0/1 loss, as this is only the case for $\alpha \rightarrow \infty$, but it tends to be very close.

\begin{figure}[t]
\centering
\begin{subfigure}[c]{0.32\textwidth}
\includegraphics[width=\textwidth,trim={0.2cm 0 0.25cm 0},clip,page=14]{binomialPWPlots.pdf}
\end{subfigure}
\begin{subfigure}[c]{0.32\textwidth}
\includegraphics[width=\textwidth,trim={0.2cm 0 0.25cm 0},clip,page=5]{binomialPWPlots.pdf}
\end{subfigure}
\begin{subfigure}[c]{0.32\textwidth}
\includegraphics[width=\textwidth,trim={0.2cm 0 0.25cm 0},clip,page=11]{binomialPWPlots.pdf}
\end{subfigure}

\begin{subfigure}[c]{0.32\textwidth}
\includegraphics[width=\textwidth,trim={0.2cm 0 0.25cm 0},clip,page=27]{binomialPWPlots.pdf}
\end{subfigure}
\begin{subfigure}[c]{0.32\textwidth}
\includegraphics[width=\textwidth,trim={0.2cm 0 0.25cm 0},clip,page=2]{binomialPWPlots.pdf}
\end{subfigure}
\begin{subfigure}[c]{0.32\textwidth}
\includegraphics[width=\textwidth,trim={0.2cm 0 0.25cm 0},clip,page=16]{binomialPWPlots.pdf}
\end{subfigure}

%\begin{subfigure}[c]{\textwidth}
\includegraphics[width=0.9\textwidth]{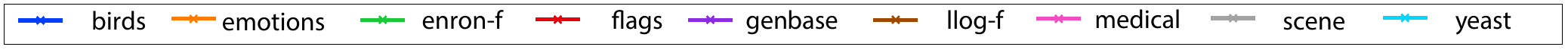}
%\end{subfigure}

\caption{Pairwise comparison of multi-label classifiers for the binomial loss.}
\label{fig:pairwise-binomial}
\end{figure}

We start the analysis with a comparison of the evaluated algorithms for the llog-f dataset.
%While the binomial loss is depicted in the upper two figures, the comparison with respect to the polynomial loss is shown at the bottom.
The graphs in Fig.~\ref{fig:datasetwise} plot the value of the parameter $k$ respectively $\alpha$ on the $x$-axis against the loss of the method on the $y$-axis.
On closer examination, we can observe some algorithms to work better than other methods for a small $k$ or $\alpha$, while the order may change as the parameter values increase and the losses demand more dependence-awareness.
For example, we can observe that PCT performs favourably to LP for small $k$ or $\alpha$, but LP catches up with increasing parameter values until it finally outperforms PCT. In general, the dependence-awareness of a learner is reflected by the slope of the performance curve (the flatter the better).

While the parameter of $\ell_{bin}$ has a simpler interpretation, as $k$ corresponds to the number of labels that is required to be predicted correctly, $\alpha$ allows for a more fine-grained analysis of dependence-awareness.  

However, with both families, we can observe intersections between the loss curves of the algorithms, explicitly showing when the order of the methods changes.

\begin{figure}[t]
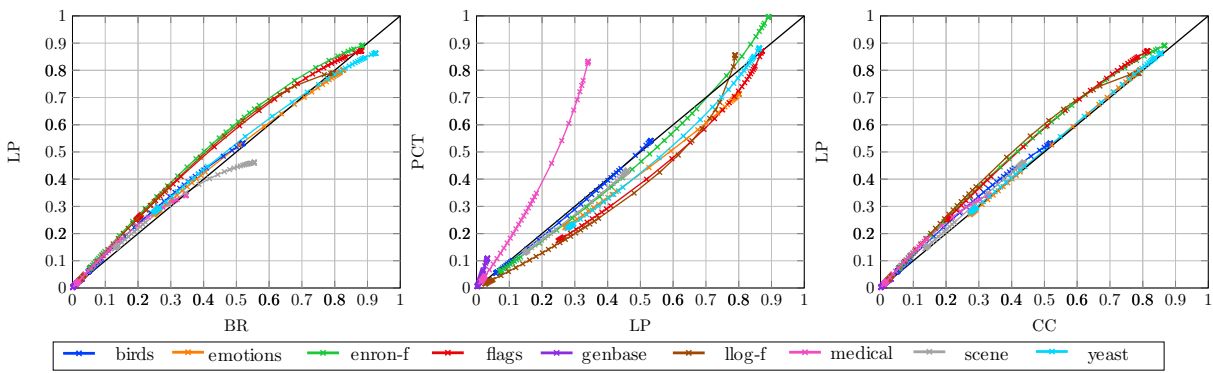

\centering
\begin{subfigure}[c]{0.32\textwidth}
\includegraphics[width=\textwidth,trim={0.2cm 0 0.25cm 0},clip,page=14]{polynomialPWPlots.pdf}
\end{subfigure}
\begin{subfigure}[c]{0.32\textwidth}
\includegraphics[width=\textwidth,trim={0.2cm 0 0.25cm 0},clip,page=5]{polynomialPWPlots.pdf}
\end{subfigure}
\begin{subfigure}[c]{0.32\textwidth}
\includegraphics[width=\textwidth,trim={0.2cm 0 0.25cm 0},clip,page=11]{polynomialPWPlots.pdf}
\end{subfigure}

% \begin{subfigure}[c]{0.32\textwidth}
% \includegraphics[width=\textwidth,trim={0.2cm 0 0.25cm 0},clip,page=17]{polynomialPWPlots.pdf}
% \end{subfigure}
% \begin{subfigure}[c]{0.32\textwidth}
% \includegraphics[width=\textwidth,trim={0.2cm 0 0.25cm 0},clip,page=2]{polynomialPWPlots.pdf}
% \end{subfigure}
% \begin{subfigure}[c]{0.32\textwidth}
% \includegraphics[width=\textwidth,trim={0.2cm 0 0.25cm 0},clip,page=16]{polynomialPWPlots.pdf}
% \end{subfigure}

%\begin{subfigure}[c]{\textwidth}
\includegraphics[width=0.9\textwidth]{legend_datasets.pdf}
%\end{subfigure}
\caption{Pairwise comparison of multi-label classifiers for the polynomial loss.}
\label{fig:pairwise-polynomial}
\end{figure}

%For a more detailed comparison of two methods, in Fig.~\ref{fig:pairwise-binomial} and Fig~\ref{fig:pairwise-polynomial}, pairwise comparisons of BR, LP, and CC as well as PCT, EPCT and RAkEL-3 are shown.
The visualizations chosen in Fig.\ \ref{fig:pairwise-binomial} and Fig.\ \ref{fig:pairwise-polynomial} allow for a more focused comparison between two methods over several datasets.
The graphs shown  (one per dataset) are produced by plotting the loss of the first learner %$l_1$ 
(on the $x$-axis) against the loss of the second learner %$l_2$ 
(on the $y$-axis) in the comparison, again varying the values of the parameters  ($1 \leq k \leq K$) for $\ell_{bin}$ and ($1 \leq \alpha \leq 1000$) for $\ell_{poly}$. 
To interpret these plots, let us highlight the following properties:
\enlargethispage*{12pt}
\begin{itemize}
\item 
Since the loss increases with increasing dependence-awareness, the direction of the graphs is from the lower left to the upper right. 
\item A point on the graph above the diagonal indicates better performance of the first method, a point below just the opposite. Thus, the intersections of the curve with the diagonal are of particular interest.
\item Also interesting is the curvature of the graph: A convex (concave) shape indicates better dependence-awareness of the first (second) method, as it improves relative to the second (first) method with increasing dependence-awareness.
\end{itemize}

Despite the different appearance in  Fig.~\ref{fig:datasetwise}, the trajectories in the pairwise comparisons are quite comparable for the two loss functions (as can be seen for the first three comparisons, respectively), demonstrating the consistency between the two losses.
In general, the experimental results confirm our expectations: 
%With an increasing dependence-awareness of the loss (increasing $k$ respectively $\alpha$), simple methods such as BR tend to perform worse than more sophisticated methods like PCT, which are able to take label dependence into account. 
With an increasing dependence-awareness of the loss (increasing $k$ respectively $\alpha$), simple methods such as BR tend to perform worse than 
dependence-aware methods like LP, which is also shown by the late crossing of the diagonal by the graphs. 
This observation is confirmed by the comparison of LP with PCT. However, compared to the case of BR, the differences at intermediate levels of dependence-awareness are larger, suggesting that PCT is better able to take label dependencies into account than BR. 
The advantage for intermediate levels is diminished if we compare to CC, a method which is less extreme than LP in its attempt to correctly predict the entire label combination. 

In contrast, RAkEL allows a more fine-grained control over the dependence-awareness with its parameter $k$, which is reflected in the comparisons in Fig.\ \ref{fig:pairwise-binomial}. 
When the ensemble members are trained to predict label subsets of size 2, RAkEL behaves quite similarly to BR, whereas for subsets of size 5 it approaches LP.
%The comparison between the two instantiation of RAkEL reveal that 
The full set of pairwise comparisons are depicted in the Fig.\ \ref{fig:pairwise-binomial-1}--\ref{fig:pairwise-polynomial-3} in the supplement.

%However, only by observing the binomial loss function family we can observe that the larger $k$ becomes, the smaller the effect the increased size of the size of labelsets on the loss function.

%To sum up, the proposed class of loss functions represent a suitable tool for a detailed analysis of multi-label algorithms. It enables insights into dependencies between labels and detailed comparisons of multi-label algorithms to reveal limitations to predict certain sizes of label subsets correctly.

%%% directly included for camera-ready version

\section{Conclusion and Future Work}
We consider a multi-label loss function as ``dependence-aware'' if it puts emphasis on getting larger label combinations right in their entirety, instead of ``merely'' making correct predictions on individual labels. In this paper, we introduced a flexible class of loss functions that allows for modeling dependence-awareness by means of non-additive measures. More specifically, we define a loss function in terms of a Choquet integral of label-wise correctness with respect to such a measure. We also proposed two instantiations of our family, in which dependence-awareness can be controlled by a single parameter, thereby ``interpolating'' between Hamming and subset 0/1 loss.

%for multi-label classification based on non-additive measures with the aim of capturing the correctness of subsets of labels. Furthermore, we presented two instantiations, the binomial and the polynomial loss, of this class, both of them exposing a single parameter to interpolate between Hamming and subset 0/1 loss. By means of an empirical study we demonstrated the two instantiations to be useful tools for the analysis of multi-label algorithms. Furthermore, despite these two instantiations being of a different shape, we showed these losses to be consistent to each other, yielding the same results when comparing multi-label algorithms in a pairwise manner.

A first experimental study has shown the potential usefulness of our loss functions as a tool for analyzing the dependence-awareness of different MLC methods, i.e., their ability to capture label dependence. Going beyond the analysis of existing algorithms, the natural next step is to develop new algorithms that are specifically tailored to our family of losses and can be customized for minimizing specific instantiations thereof. 

%Regarding the capabilities of multi-label algorithms getting label subsets of a certain size correctly, first insights could already be gained. Using the proposed class of loss functions as a tool, an expansion of the experiments and the analysis to better understand better the way multi-label algorithms perform outlines interesting future work. 

\section*{Acknowledgement}
This work was partially supported by the German Research Foundation (DFG) within the Collaborative Research Center ``On-The-Fly Computing'' (SFB 901/3, grant no.\ 160364472) and the DFG project ``Multilabel Rule Learning'' under grant no.\ 400845550.
The authors also gratefully acknowledge support of this project through computing time provided by the Paderborn Center for Parallel Computing (PC$^2$).

\bibliographystyle{abbrvnat}
\bibliography{literature}

\newpage
\appendix

\pagebreak
\section{Pairwise Comparisons of Learners Regarding $\ell_{bin}$}
\begin{figure}[ht!]
\centering
\includegraphics[width=0.32\textwidth,trim={0.2cm 0 0.25cm 0},clip,page=1]{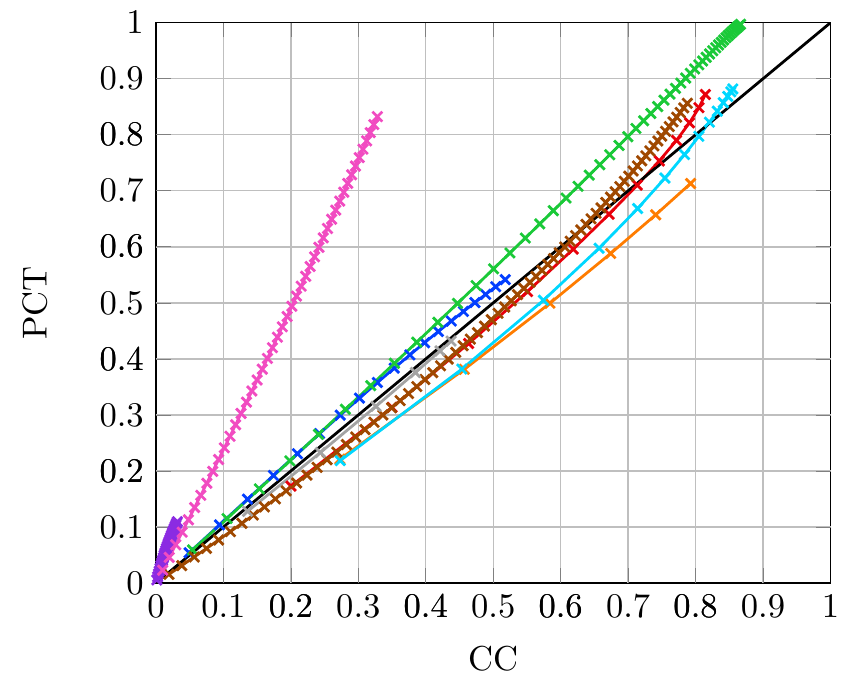}
\includegraphics[width=0.32\textwidth,trim={0.2cm 0 0.25cm 0},clip,page=2]{binomialPWPlots.pdf}
\includegraphics[width=0.32\textwidth,trim={0.2cm 0 0.25cm 0},clip,page=3]{binomialPWPlots.pdf}

\includegraphics[width=0.32\textwidth,trim={0.2cm 0 0.25cm 0},clip,page=4]{binomialPWPlots.pdf}
\includegraphics[width=0.32\textwidth,trim={0.2cm 0 0.25cm 0},clip,page=5]{binomialPWPlots.pdf}
\includegraphics[width=0.32\textwidth,trim={0.2cm 0 0.25cm 0},clip,page=6]{binomialPWPlots.pdf}

\includegraphics[width=0.32\textwidth,trim={0.2cm 0 0.25cm 0},clip,page=7]{binomialPWPlots.pdf}
\includegraphics[width=0.32\textwidth,trim={0.2cm 0 0.25cm 0},clip,page=8]{binomialPWPlots.pdf}
\includegraphics[width=0.32\textwidth,trim={0.2cm 0 0.25cm 0},clip,page=9]{binomialPWPlots.pdf}

\includegraphics[width=0.32\textwidth,trim={0.2cm 0 0.25cm 0},clip,page=10]{binomialPWPlots.pdf}
\includegraphics[width=0.32\textwidth,trim={0.2cm 0 0.25cm 0},clip,page=11]{binomialPWPlots.pdf}
\includegraphics[width=0.32\textwidth,trim={0.2cm 0 0.25cm 0},clip,page=12]{binomialPWPlots.pdf}

\includegraphics[width=.875\textwidth]{legend_datasets.pdf}

\caption{Pairwise comparison of multi-label classifiers for the binomial loss (part 1).}
\label{fig:pairwise-binomial-1}
\end{figure}

\begin{figure*}[h]
\centering
\includegraphics[width=0.32\textwidth,trim={0.2cm 0 0.25cm 0},clip,page=13]{binomialPWPlots.pdf}
\includegraphics[width=0.32\textwidth,trim={0.2cm 0 0.25cm 0},clip,page=14]{binomialPWPlots.pdf}
\includegraphics[width=0.32\textwidth,trim={0.2cm 0 0.25cm 0},clip,page=15]{binomialPWPlots.pdf}

\includegraphics[width=0.32\textwidth,trim={0.2cm 0 0.25cm 0},clip,page=16]{binomialPWPlots.pdf}
\includegraphics[width=0.32\textwidth,trim={0.2cm 0 0.25cm 0},clip,page=17]{binomialPWPlots.pdf}
\includegraphics[width=0.32\textwidth,trim={0.2cm 0 0.25cm 0},clip,page=18]{binomialPWPlots.pdf}

\includegraphics[width=0.32\textwidth,trim={0.2cm 0 0.25cm 0},clip,page=19]{binomialPWPlots.pdf}
\includegraphics[width=0.32\textwidth,trim={0.2cm 0 0.25cm 0},clip,page=20]{binomialPWPlots.pdf}
\includegraphics[width=0.32\textwidth,trim={0.2cm 0 0.25cm 0},clip,page=21]{binomialPWPlots.pdf}

\includegraphics[width=0.32\textwidth,trim={0.2cm 0 0.25cm 0},clip,page=22]{binomialPWPlots.pdf}
\includegraphics[width=0.32\textwidth,trim={0.2cm 0 0.25cm 0},clip,page=23]{binomialPWPlots.pdf}
\includegraphics[width=0.32\textwidth,trim={0.2cm 0 0.25cm 0},clip,page=25]{binomialPWPlots.pdf}

\includegraphics[width=.875\textwidth]{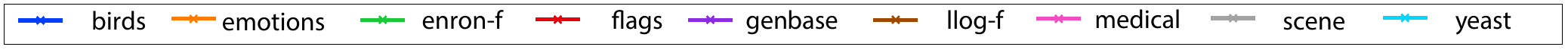}

\caption{Pairwise comparison of multi-label classifiers for the binomial loss (part 2).}
\label{fig:pairwise-binomial-2}
\end{figure*}

\begin{figure*}[h]
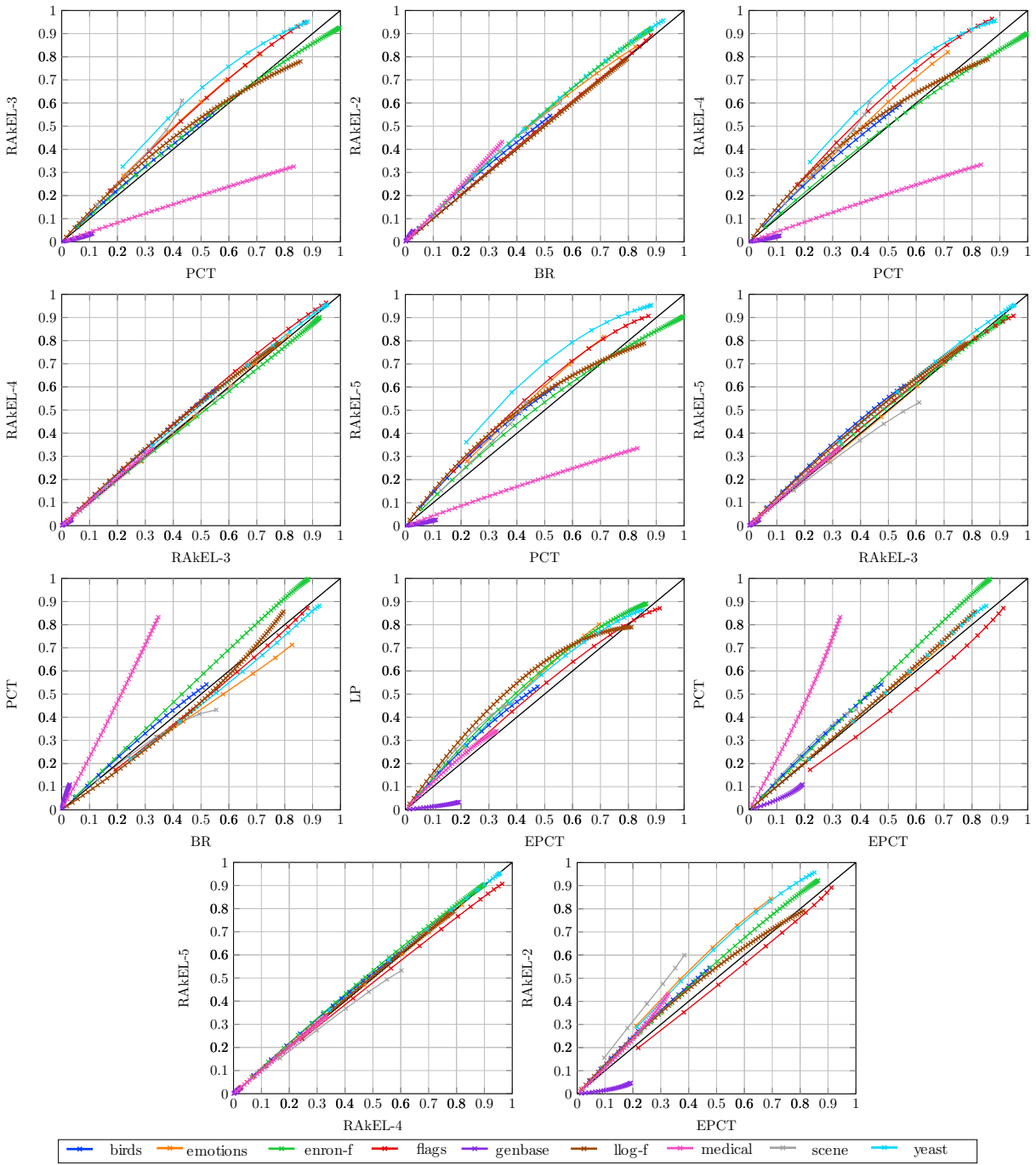

\centering
\includegraphics[width=0.32\textwidth,trim={0.2cm 0 0.25cm 0},clip,page=26]{binomialPWPlots.pdf}
\includegraphics[width=0.32\textwidth,trim={0.2cm 0 0.25cm 0},clip,page=27]{binomialPWPlots.pdf}
\includegraphics[width=0.32\textwidth,trim={0.2cm 0 0.25cm 0},clip,page=28]{binomialPWPlots.pdf}

\includegraphics[width=0.32\textwidth,trim={0.2cm 0 0.25cm 0},clip,page=29]{binomialPWPlots.pdf}
\includegraphics[width=0.32\textwidth,trim={0.2cm 0 0.25cm 0},clip,page=30]{binomialPWPlots.pdf}
\includegraphics[width=0.32\textwidth,trim={0.2cm 0 0.25cm 0},clip,page=31]{binomialPWPlots.pdf}

\includegraphics[width=0.32\textwidth,trim={0.2cm 0 0.25cm 0},clip,page=32]{binomialPWPlots.pdf}
\includegraphics[width=0.32\textwidth,trim={0.2cm 0 0.25cm 0},clip,page=33]{binomialPWPlots.pdf}
\includegraphics[width=0.32\textwidth,trim={0.2cm 0 0.25cm 0},clip,page=34]{binomialPWPlots.pdf}

\includegraphics[width=0.32\textwidth,trim={0.2cm 0 0.25cm 0},clip,page=35]{binomialPWPlots.pdf}
\includegraphics[width=0.32\textwidth,trim={0.2cm 0 0.25cm 0},clip,page=36]{binomialPWPlots.pdf}

\includegraphics[width=.875\textwidth]{legend_datasets.pdf}

\caption{Pairwise comparison of multi-label classifiers for the binomial loss (part 3).}
\label{fig:pairwise-binomial-3}
\end{figure*}

\FloatBarrier

\section{Pairwise Comparisons of Learners Regarding $\ell_{poly}$}
\begin{figure}[ht!]
\centering
\includegraphics[width=0.32\textwidth,trim={0.2cm 0 0.25cm 0},clip,page=1]{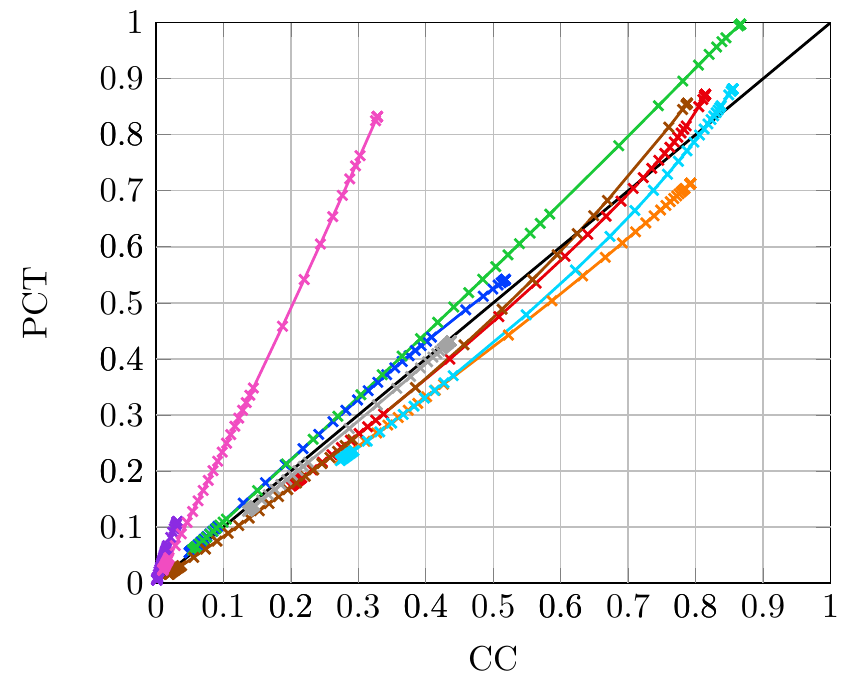}
\includegraphics[width=0.32\textwidth,trim={0.2cm 0 0.25cm 0},clip,page=2]{polynomialPWPlots.pdf}
\includegraphics[width=0.32\textwidth,trim={0.2cm 0 0.25cm 0},clip,page=3]{polynomialPWPlots.pdf}

\includegraphics[width=0.32\textwidth,trim={0.2cm 0 0.25cm 0},clip,page=4]{polynomialPWPlots.pdf}
\includegraphics[width=0.32\textwidth,trim={0.2cm 0 0.25cm 0},clip,page=5]{polynomialPWPlots.pdf}
\includegraphics[width=0.32\textwidth,trim={0.2cm 0 0.25cm 0},clip,page=6]{polynomialPWPlots.pdf}

\includegraphics[width=0.32\textwidth,trim={0.2cm 0 0.25cm 0},clip,page=7]{polynomialPWPlots.pdf}
\includegraphics[width=0.32\textwidth,trim={0.2cm 0 0.25cm 0},clip,page=8]{polynomialPWPlots.pdf}
\includegraphics[width=0.32\textwidth,trim={0.2cm 0 0.25cm 0},clip,page=9]{polynomialPWPlots.pdf}

\includegraphics[width=0.32\textwidth,trim={0.2cm 0 0.25cm 0},clip,page=10]{polynomialPWPlots.pdf}
\includegraphics[width=0.32\textwidth,trim={0.2cm 0 0.25cm 0},clip,page=11]{polynomialPWPlots.pdf}
\includegraphics[width=0.32\textwidth,trim={0.2cm 0 0.25cm 0},clip,page=12]{polynomialPWPlots.pdf}

\includegraphics[width=.875\textwidth]{legend_datasets.pdf}

\caption{Pairwise comparison of multi-label classifiers for the polynomial loss (part 1).}
\label{fig:pairwise-polynomial-1}
\end{figure}

\begin{figure*}[h]
\centering
\includegraphics[width=0.32\textwidth,trim={0.2cm 0 0.25cm 0},clip,page=13]{polynomialPWPlots.pdf}
\includegraphics[width=0.32\textwidth,trim={0.2cm 0 0.25cm 0},clip,page=14]{polynomialPWPlots.pdf}
\includegraphics[width=0.32\textwidth,trim={0.2cm 0 0.25cm 0},clip,page=15]{polynomialPWPlots.pdf}

\includegraphics[width=0.32\textwidth,trim={0.2cm 0 0.25cm 0},clip,page=16]{polynomialPWPlots.pdf}
\includegraphics[width=0.32\textwidth,trim={0.2cm 0 0.25cm 0},clip,page=17]{polynomialPWPlots.pdf}
\includegraphics[width=0.32\textwidth,trim={0.2cm 0 0.25cm 0},clip,page=18]{polynomialPWPlots.pdf}

\includegraphics[width=0.32\textwidth,trim={0.2cm 0 0.25cm 0},clip,page=19]{polynomialPWPlots.pdf}
\includegraphics[width=0.32\textwidth,trim={0.2cm 0 0.25cm 0},clip,page=20]{polynomialPWPlots.pdf}
\includegraphics[width=0.32\textwidth,trim={0.2cm 0 0.25cm 0},clip,page=21]{polynomialPWPlots.pdf}

\includegraphics[width=0.32\textwidth,trim={0.2cm 0 0.25cm 0},clip,page=22]{polynomialPWPlots.pdf}
\includegraphics[width=0.32\textwidth,trim={0.2cm 0 0.25cm 0},clip,page=23]{polynomialPWPlots.pdf}
\includegraphics[width=0.32\textwidth,trim={0.2cm 0 0.25cm 0},clip,page=25]{polynomialPWPlots.pdf}

\includegraphics[width=.875\textwidth]{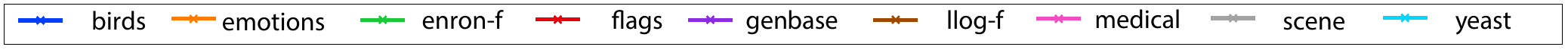}

\caption{Pairwise comparison of multi-label classifiers for the polynomial loss (part 2).}
\label{fig:pairwise-polynomial-2}
\end{figure*}

\begin{figure*}[h]
\centering
\includegraphics[width=0.32\textwidth,trim={0.2cm 0 0.25cm 0},clip,page=26]{polynomialPWPlots.pdf}
\includegraphics[width=0.32\textwidth,trim={0.2cm 0 0.25cm 0},clip,page=27]{polynomialPWPlots.pdf}
\includegraphics[width=0.32\textwidth,trim={0.2cm 0 0.25cm 0},clip,page=28]{polynomialPWPlots.pdf}

\includegraphics[width=0.32\textwidth,trim={0.2cm 0 0.25cm 0},clip,page=29]{polynomialPWPlots.pdf}
\includegraphics[width=0.32\textwidth,trim={0.2cm 0 0.25cm 0},clip,page=30]{polynomialPWPlots.pdf}
\includegraphics[width=0.32\textwidth,trim={0.2cm 0 0.25cm 0},clip,page=31]{polynomialPWPlots.pdf}

\includegraphics[width=0.32\textwidth,trim={0.2cm 0 0.25cm 0},clip,page=32]{polynomialPWPlots.pdf}
\includegraphics[width=0.32\textwidth,trim={0.2cm 0 0.25cm 0},clip,page=33]{polynomialPWPlots.pdf}
\includegraphics[width=0.32\textwidth,trim={0.2cm 0 0.25cm 0},clip,page=34]{polynomialPWPlots.pdf}

\includegraphics[width=0.32\textwidth,trim={0.2cm 0 0.25cm 0},clip,page=35]{polynomialPWPlots.pdf}
\includegraphics[width=0.32\textwidth,trim={0.2cm 0 0.25cm 0},clip,page=36]{polynomialPWPlots.pdf}

\includegraphics[width=.875\textwidth]{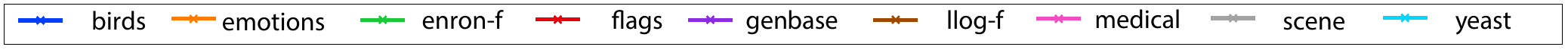}

\caption{Pairwise comparison of multi-label classifiers for the polynomial loss (part 3).}
\label{fig:pairwise-polynomial-3}
\end{figure*}
\end{document}